\documentclass[10pt, a4paper]{article}
\usepackage{lrec2022} % this is the new LREC2022 Style
\usepackage{multibib}
\newcites{languageresource}{Language Resources}
\usepackage{graphicx}
\usepackage{tabularx}
\usepackage{soul}
\usepackage{titlesec}
\titleformat{\section}{\normalfont\large\bfseries\center}{\thesection.}{1em}{}
\titleformat{\subsection}{\normalfont\SmallTitleFont\bfseries\raggedright}{\thesubsection.}{1em}{}
\titleformat{\subsubsection}{\normalfont\normalsize\bfseries\raggedright}{\thesubsubsection.}{1em}{}
\renewcommand\thesection{\arabic{section}}
\renewcommand\thesubsection{\thesection.\arabic{subsection}}
\renewcommand\thesubsubsection{\thesubsection.\arabic{subsubsection}}
\usepackage{epstopdf}
\usepackage[utf8]{inputenc}

\usepackage[hidelinks]{hyperref}
\usepackage{xstring}

\usepackage{color}

\usepackage{pgfplots, tikz}
\pgfplotsset{compat=1.18}

\usepackage{booktabs}
\usepackage{adjustbox}
\usepackage[para]{threeparttable}
\usepackage{lipsum}
\usepackage{float}
\usepackage{multirow}

\title{Comparison of Conventional Hybrid and CTC/Attention Decoders for Continuous Visual Speech Recognition}

\name{David Gimeno-Gómez, Carlos-D. Martínez-Hinarejos} 

\address{Pattern Recognition and Human Language Technologies Research Center\\
  Universitat Politècnica de València, Camino de Vera, s/n, 46022, València, Spain \\
         {\tt dagigo1@dsic.upv.es, cmartine@dsic.upv.es}         
}

\abstract{
Thanks to the rise of deep learning and the availability of large-scale audio-visual databases, recent advances have been achieved in Visual Speech Recognition (VSR). Similar to other speech processing tasks, these end-to-end VSR systems are usually based on encoder-decoder architectures. While encoders are somewhat general, multiple decoding approaches have been explored, such as the conventional hybrid model based on Deep Neural Networks combined with Hidden Markov Models (DNN-HMM) or the Connectionist Temporal Classification (CTC) paradigm. However, there are languages and tasks in which data is scarce, and in this situation, there is not a clear comparison between different types of decoders. Therefore, we focused our study on how the conventional DNN-HMM decoder and its state-of-the-art CTC/Attention counterpart behave depending on the amount of data used for their estimation. We also analyzed to what extent our visual speech features were able to adapt to scenarios for which they were not explicitly trained, either considering a similar dataset or another collected for a different language. Results showed that the conventional paradigm reached recognition rates that improve the CTC/Attention model in data-scarcity scenarios along with a reduced training time and fewer parameters.
 \\ \newline  
}

\begin{document}

\maketitleabstract

\section{Introduction}
\label{sec:introduction}
% -- What is VSR?
Inspired by different studies that have shown the relevance of visual cues during our speech perception process \cite{mcgurk1976hearing,besle2004bimodal}, several authors explored the task of Automatic Speech Recognition (ASR) from an audio-visual perspective \cite{potamianos2003recent,afouras2018deep,maja2021conformers}. These works supported that auditory and visual cues complement each other, leading to more robust systems, especially in adverse scenarios such as a noisy environment \cite{juang1991adverse}.

Nonetheless, in the last few decades, there has been an increasing interest in Visual Speech Recognition (VSR) \cite{fernandez2018survey}, a challenging task that aims to interpret speech solely by reading the speaker's lips. Recognizing speech without the need for the auditory sense can offer a wide range of applications, e.g., silent visual passwords \cite{silent2020passwd}, active speaker detection \cite{kim2021asd,tao2021talknetasd}, visual keyword spotting \cite{stafylakis2018zero,prajwal2021vks}, or the development of silent speech interfaces that would be able to improve the lives of people who experience difficulties in producing speech \cite{denby2010silent,ssi2020review}.

Although unprecedented results have recently been achieved in the field \cite{ma2022visual,shi2022learning,prajwal2021sub}, VSR remains an open research problem, where different factors must be considered, e.g., visual ambiguities~\cite{bear2014phoneme,fernandez2017optimizing}, the complex modeling of silence~\cite{thangthai2018computer}, the inter-personal variability among speakers~\cite{cox2008challenge}, and different lighting conditions, as well as more technical aspects such as frame rate and image resolution~\cite{bear2016decoding,bear2014resolution,dungan2018impact}.

Motivated by these challenges, one of the main research purposes in the field of VSR has been exploring diverse approaches to extract powerful visual speech representations \cite{fernandez2018survey}. Traditional techniques, either based on Principal Component Analysis (PCA), Discrete Cosine Transform (DCT), or Active Shape Models (AAM), were the object of study for decades \cite{bregler94eigenlips,matthews2002extraction,potamianos2003recent,gimenogomez21iberspeech}. Nowadays, the most common approach is the design of end-to-end architectures \cite{ma2022visual,shi2022learning,prajwal2021sub}, models capable of automatically learning these visual speech representations in a data-driven manner. However, to the best of our knowledge, there is no systematic analysis of how robust these data-driven features are to domain- and language-mismatch scenarios.

% In this work, we conducted a systematic study to assess the robustness of these data-driven features.

Regarding visual speech decoders, different approaches have been explored in the literature. From the systems based on Hidden Markov Models combined with Gaussian Mixture Models (GMM-HMM) \cite{juang1991hidden,gales2008application}, the use of Deep Neural Networks (DNNs) to model emission probabilities provided the so-called hybrid DNN-HMM model \cite{hinton2012deep}. However, these conventional systems present limitations \cite{watanabe2017ctcattention} because of the need for forced alignments, a pre-defined lexicon, independent module optimizations, and the use of pre-defined, non-adaptive visual speech representations. Hence, research shifted towards end-to-end architectures based on Deep Learning techniques \cite{asr2019survey}. Early works, either based on Recurrent Neural Networks (RNNs) \cite{chan2016listen} or on the Connectionist Temporal Classification (CTC) paradigm \cite{graves2006connectionist,graves2014towards} demonstrated what these novel architectures were capable of. Thereafter, based on advances in neural machine translation \cite{vaswani2017attention}, remarkable results were obtained thanks to the use of powerful attention-based mechanisms \cite{speech2018transformer}. Nowadays, the hybrid CTC/Attention decoder, which combines the properties of both paradigms, is considered the current state of the art in speech processing \cite{watanabe2017ctcattention,maja2021conformers}.
% Besides, an RNN-based Transducer \cite{graves2012sequence,streaming2019e2e} was proposed to address streaming ASR specifically.

Most comparative studies on decoding paradigms were conducted for auditory-based ASR \cite{luscher19interspeech,watanabe2019comparison,afouras2018deep}. However, albeit different speech decoders were explored in VSR \cite{thangthai2017,chung2017lrs,afouras2018deep,ma2022visual}, it is not easy to compare all these approaches due to the use of different databases or the design of different experimental setups \cite{fernandez2018survey}. 

These were the main reasons that motivated our research, where our key contributions are: 
\begin{itemize}
    \item A comprehensive comparison of conventional hybrid DNN-HMM decoders and their state-of-the-art CTC/Attention counterpart for the continuous VSR task.

    \item We systematically studied how these different decoding paradigms behave based on the amount of data available for their estimation, showing that conventional HMM-based systems significantly outperformed state-of-the-art architectures in data-scarcity scenarios.

    \item We discussed different deployment aspects, such as training time, number of parameters, or real-time factor, supporting a more appropriate model selection where not only performance is considered.

    \item We analyzed to what extent our pre-trained data-driven visual speech features were able to adapt to scenarios for which they were not explicitly trained by considering databases covering a different domain or language.
    
\end{itemize}

\section{Related Work}
\label{sec:related-work}
\noindent\textbf{Visual Speech Features.} Unlike auditory-based ASR, there was no consensus on the most suitable visual speech representation for decades \cite{fernandez2018survey}. In the past, diverse traditional techniques were widely explored \cite{bregler94eigenlips,matthews2002extraction,potamianos2003recent,gimenogomez21iberspeech}. Nowadays, the design of end-to-end architectures, capable of automatically learning these speech representations in a data-driven manner during their training process, has led to unprecedented advances in the field \cite{ma2022visual,shi2022learning,prajwal2021sub}. Most of these approaches rely on convolutional neural networks, such as the so-called ResNet \cite{he2016resnet}, to obtain a latent visual representation, and then, attention-based mechanisms \cite{vaswani2017attention} are used to model temporal relationships. In addition, self-supervised methods, complemented with acoustic cues during their estimation, have also been explored \cite{ma2021lira,shi2022learning}. However, there are no studies on how these data-driven features can adapt or transfer their knowledge when dealing with different domains or languages which they were not explicitly trained for.

\noindent\textbf{Visual Speech Recognition.} Albeit conventional paradigms were explored for VSR \cite{thangthai2017,lrec2022liprtve}, the current state of the art is dominated by end-to-end approaches based on powerful attention-mechanisms \cite{ma2022visual,shi2022learning,prajwal2021sub}. On average, results of around 25-30\% Word Error Rate (WER) were achieved for the English corpora LRS2-BBC \cite{afouras2018deep} and LRS3-TED \cite{afouras2018lrs3}. However, by the use of large-scale pseudo-label \cite{ma2023autoavsr} or synthetic data \cite{liu2023synthvsr}, recent works have reached a new state of the art around 15\% WER. 

Besides, interest in VSR for languages other than English has recently increased \cite{ma2022visual,lrec2022liprtve}. A remarkable work in this regard is the one carried out by \newcite{anwar2023muavic}, where 8 non-English languages were explored presenting a new multi-lingual benchmark. However, the authors focused on audio-visual speech recognition/translation and did not report results for lipreading. Regarding Spanish VSR (a language also considered in our work), \newcite{ma2022visual} reached recognition rates of around 50\% WER for different Spanish corpora, using an end-to-end architecture, while \newcite{lrec2022liprtve} presented the challenging LIP-RTVE database, reporting baseline results (roughly 95\% WER) using traditional visual speech features and a GMM-HMM model. Subsequently, although results around 60\% WER were achieved for the LIP-RTVE corpus, the same authors focused their study on speaker-dependent scenarios \cite{gimeno2023comparing}.

Due to the use of different databases, model architectures, and experimental setups, it is not easy to adequately compare all approaches explored in the literature \cite{fernandez2018survey,nemani2023survey}.

\noindent\textbf{Speech Decoders:} Multiple studies in model comparison for auditory-based ASR have been developed. \newcite{luscher19interspeech} presented a comparison between a conventional DNN-HMM model and an end-to-end RNN-based architecture. Their results showed that the DNN-HMM paradigm outperformed the end-to-end recognizer. \newcite{watanabe2019comparison} carried out a thorough comparative study focused on RNN- and Transformed-based end-to-end models, including HMM-based models in the comparison. Although their findings showed that Transformer-based models outperformed RNNs, they also demonstrated that a DNN-HMM model could reach state-of-the-art recognition rates. 

Regarding VSR, to the best of our knowledge, \cite{afouras2018deep} is the only work that includes a systems comparison. Specifically, it compared the CTC paradigm to the attention-based one, showing that, albeit each one offers different valuable properties, the attention-based approach performed better, probably due to its powerful context modeling. However, although databases of different natures and multiple architectures were explored in the literature, none of the previous works conducted a systematic study on how these different decoding paradigms behave for VSR, depending on the data available for training. 

\noindent \textbf{Present Work.} Motivated by all these aspects, we present a comprehensive comparison of conventional hybrid DNN-HMM decoders and their state-of-the-art CTC/Attention counterpart for the continuous VSR task. We not only systematically compared both approaches based on the amount of data available for their estimation, but we also took into account different deployment aspects. In addition,  by considering three benchmarking VSR datasets, we evaluated how robust our pre-trained data-driven visual speech features could be to domain- and language-mismatch scenarios.

\section{Method}
\label{sec:method}
% The purpose of our research is not only to compare two different decoding paradigms, but also to identify which of them is more appropriate depending on the amount of data that we have when addressing a specific VSR task. Therefore, a comparative study in terms of system performance and training time is proposed. Specifically, we analyzed how the conventional hybrid DNN-HMM decoder \cite{hinton2012deep} and its state-of-the-art CTC/Attention counterpart \cite{watanabe2017ctcattention} performed based on the number of hours they were trained with. By considering several databases, we also investigated to what extent the pre-trained visual speech encoder was able to adapt to scenarios for which it was not explicitly trained.

\subsection{Databases} \label{sec:databases}

\noindent\textbf{LRS2-BBC} \cite{afouras2018deep} is a large-scale English database composed of around 224 hours collected from BBC TV programs. It consists of a pre-training set with 96,318 samples (195 hours), a training set with 45,839 samples (28 hours), a validation set with 1,082 samples (0.6 hours), and a test set with 1,243 samples (0.5 hours). It offers more than 2 million running words with a vocabulary size of around 40k different words. This corpus represents our ideal scenario, since our visual speech encoder was explicitly trained for this task.

\noindent\textbf{LRS3-TED} \cite{afouras2018lrs3} is the largest publicly audio-visual English database offering around 438 hours. It was collected from TED talks, consisting of a `pre-train' set with 118,516 samples (407 hours), a `train-val' set with 31,982 samples (30 hours), and a test set with 1,321 samples (0.9 hours).  It comprises more than 4 million running words with a vocabulary size of around 50k different words. This corpus represents our domain-mismatch scenario.

\noindent\textbf{LIP-RTVE} \cite{lrec2022liprtve} is a challenging Spanish database collected from TV newscast programs, providing around 13 hours of data. Its speaker-independent partition consists of a training set with 7,142 samples (9 hours), a validation set with 1638 samples (2 hours), and a test set with 1572 samples (2 hours). It provides more than 100k running words with a vocabulary size of around 10k different words. This corpus represents our language-mismatch scenario.

\subsection{Visual Speech Encoder} \label{sec:visual-encoder}

\begin{figure*}[htbp]
  \centering
  \includegraphics[width=0.8\textwidth]{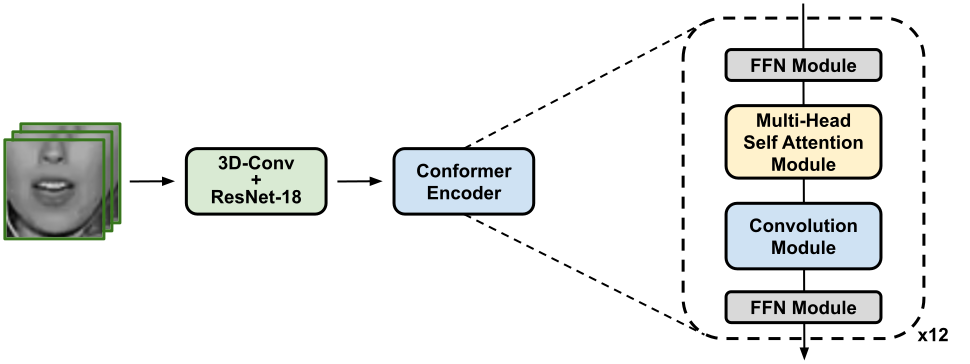}
  \caption{The overall architecture of our visual speech encoder. For simplicity, the initial layer normalization, the residual connection, and the final dropout of each module that compose the conformer encoder are omitted. Conv and FFN refer to Convolutional layer and Feed Forward Network, respectively.}
  \label{fig:encoder-scheme}
\end{figure*}

A pre-trained encoder is used to extract our 256-dimensional visual speech features. As reflected in Figure \ref{fig:encoder-scheme}, the encoder is based on the state-of-the-art architecture designed by \newcite{ma2022visual}, where two different blocks are distinguished.

\noindent \textbf{Visual Frontend.} A 3D convolutional layer with a kernel size of 7x7 pixels and a receptive field of 5 frames is used to deal with spatial relationships. Once the video stream data is flattened along the temporal dimension, a 2D ResNet-18 \cite{he2016resnet} focuses on capturing local visual patterns. The Swish activation function \cite{swish2017prajit} was used. This visual frontend comprises about 11 million parameters.
% first, a relative positional embedding \cite{relpos2019transformerxl} is applied. Then, 

\noindent \textbf{Temporal Encoder.} A 12-layer Conformer encoder \cite{gulati20interspeech} is defined to capture both global and local speech interactions across time from the previous visual latent representation. Each layer is composed of four modules, namely two feed forward networks in a macaron style, a multi-head self-attention module, and a convolution module. Layer normalization precedes each module, while a residual connection and a final dropout are applied over its output. The main difference w.r.t. the original Transformer encoder architecture \cite{vaswani2017attention} is the convolution module, which is able to model local temporal speech patterns by using point- and depth-wise convolutions. This temporal encoder comprises about 32 million parameters.

\subsection{Conventional Hybrid Decoder} \label{conventional-decoder}

\noindent \textbf{Morphological Model.} The design of our conventional DNN-HMM decoder was based on the Wall Street Journal recipe\footnote{\url{https://github.com/kaldi-asr/kaldi/tree/master/egs/wsj/s5}} provided by the Kaldi toolkit \cite{povey2011kaldi}. First of all, we estimated a preliminary GMM-HMM to obtain temporal alignments. Then, we applied the so-called HiLDA technique \cite{potamianos2001hierarchical}, reducing our visual speech features to a 40-dimensional latent representation.

Regarding our best DNN-HMM architecture, it consisted of two hidden layers of 1024 units, each followed by a Sigmoid activation function. We defined as input an 11-frame context window over the previous HiLDA features. The output layer dimension depended on the number of HMM-state labels defined by the preliminary GMM-HMM. Concretely: 3624, 3304, and 1968 HMM-state labels were defined for the LRS2-BBC, LRS3-TED, and LIP-RTVE corpora, respectively. Then, we estimated our DNN-HMM system based on a frame cross-entropy training \cite{hinton2012deep}. In average terms, each DNN-HMM decoder comprised around 4 million parameters.

Another important aspect was the HMM's topology. Due to the lower sample rate presented by visual cues compared to acoustic signals, our first experiments focused on the optimal HMM's topology. By adding transitions and/or reducing the number of states, we found that, in all cases, a 3-state left-to-right topology with skip transitions to the final state was the best approach to fit the temporary nature of our visual data.
%the topology depicted in Figure \ref{fig:hmm-topology} 

\noindent \textbf{Lexicon Model.} For LRS2-BBC and LRS3-TED, we processed their corresponding training transcriptions with a phonemizer\footnote{\url{www.github.com/Kyubyong/g2p}} based on the CMU pronunciation dictionary\footnote{\url{www.speech.cs.cmu.edu/cgi-bin/cmudict}}. Similarly, for LIP-RTVE, we considered a phonemizer based on Spanish phonetic rules \cite{quilis1997principios}. However, the amount of training data of LIP-RTVE is not comparable to its English counterparts. For this reason, we used the text provided by the LIP-RTVE's authors for the estimation of a language model, which offers around 80k phrases collected from different but contemporary TV newscasts\footnote{It comprises 1.5 million running words with a vocabulary size of around 45k different words}. Thus, a set of 39 and 24 phonemes were defined for English and Spanish, respectively. In both cases, the default \emph{silence} phones of Kaldi were then included.

\noindent \textbf{Language Model.} We used a 6-layer character-level Language Model (LM) based on the Transformer architecture \cite{vaswani2017attention}. A more detailed description about this Transformer-based LM can be found in Subsection \ref{sec:impl-lm}.

However, for this conventional paradigm, we applied an approach based on a combination of one-pass decoding and lattice re-scoring \cite{luscher19interspeech}. Consequently, we used an auxiliary n-gram language model for decoding before re-scoring with the Transformed-based LM. Hence, a 3-gram word-level LM was also estimated. For each English database, we used the transcriptions included in its corresponding training set, while for LIP-RTVE, we considered the aforementioned 80k phrases. The estimated n-gram LMs offered 100.5, 112.8, and 113.4 test perplexities, with 25, 16, and 193 out-of-vocabulary words, for LRS2-BBC, LRS3-TED, and LIP-RTVE, respectively.

\noindent \textbf{Decoding.} The decoder is defined as a weighted finite-state transducer integrating the morphological, lexicon, and language models. Readers are referred to \newcite{mohri2008speech} for more details.

\subsection{CTC/Attention Decoder} \label{sec:ctc-attention}

\noindent \textbf{Morphological Model.} This state-of-the-art approach was implemented using the ESPNet toolkit \cite{watanabe18espnet}. Specifically, the decoder was composed of a 6-layer Transformer decoder \cite{vaswani2017attention} and a fully connected layer as the CTC-based decoding branch \cite{graves2006connectionist}. By combining both paradigms, the model is able to adopt both the Markov assumptions of CTC (an aspect in harmony with the speech nature) and the flexibility of the non-sequential alignments provided by the attention-based decoder. As proposed by \newcite{watanabe2017ctcattention}, the loss function is defined as follows:

\vspace{-0.4cm}
\begin{equation} \label{eq:ctc-attention-loss}
    \mathcal{L} = \alpha \log p_{ctc}(\textbf{y}|\textbf{x}) + (1 - \alpha) \log p_{attn}(\textbf{y}|\textbf{x})
\end{equation}

\noindent where $p_{ctc}$ and $p_{attn}$ denote the CTC and Attention posteriors, respectively. In both terms, $\textbf{x}$ and $\textbf{y}$ refer to the input visual stream and its corresponding character-level target, respectively.  The $\alpha$ weight balances the relative influence of each decoder.

It should be noted that this end-to-end approach is based on a character-level speech recognition. For LRS2-BBC and LRS3-TED, we considered a set of 41 characters, while for LIP-RTVE we used a set of 37 characters. In both cases, special characters were included, such as the `space' and the `blank' symbols. On average, each CTC/Attention decoder comprised around 9.5 million parameters.

\noindent \textbf{Language Model.} We used a 6-layer character-level LM based on the Transformer architecture \cite{vaswani2017attention}. More details about how it was estimated are found in Subsection \ref{sec:impl-lm}.

\noindent \textbf{Decoding.} The decoder integrates the attention- and CTC-based branches and the Transformer-based LM in a beam search process. Albeit it is the attention-based branch that leads this decoding process until predicting the end-of-sentence token, the rest of the components influence the search in a shallow fusion manner, as reflected in:

\begin{equation} \label{eq:ctc-attention-decoding}
    S = \lambda S_{ctc} + (1 - \lambda)S_{attn} + \beta S_{lm}
\end{equation}

\noindent where $S_{ctc}$ and $S_{attn}$ are the scores of the CTC and the Attention decoder, respectively, $\lambda$ is their corresponding relative weight, and $\beta$ and $S_{lm}$ refer to the LM influence weight and the LM score, respectively. Readers are referred to \newcite{watanabe2017ctcattention} for a more detailed description.

\section{Experimental Setup}
\label{sec:experiments}

\begin{figure*}[!htbp] 
  \centering
  \begin{tikzpicture}[scale=0.7]
    \begin{axis}[xlabel=\textbf{Hours of Training Data},
                 ylabel=\textbf{\% WER},
                 symbolic x coords={,1,2,5,9,10,20,50,100,200,223,437,},
                 width=\textwidth,
                 height=0.6\textwidth,
                 legend cell align={left},
                 ]

    %% LIP-RTVE %%
    \addplot[thick, mark=x, mark options={solid}, red, error bars/.cd, y dir=both, y explicit] 
      plot coordinates {
        (1,78.1) +- (0,0.9)
        (2,77.5) +- (0,1.0)
        (5,67.8) +- (0,1.1)
        (9,66.2) +- (0,1.1)
      };
      \addlegendentry{LIP-RTVE (DNN-HMM)}

    \addplot[thick, dashed, mark=x, mark options={solid}, red, error bars/.cd, y dir=both, y explicit] 
      plot coordinates {
        (1,100.0) +- (0,0.0)
        (2,100.0) +- (0,0.0)
        (5,100.0) +- (0,0.0)
        (9,89.0) +- (0,1.2)
      };
      \addlegendentry{LIP-RTVE (CTC/Attention)}

    %% LRS3-TED %%
    \addplot[thick, mark=x, mark options={solid}, blue, error bars/.cd, y dir=both, y explicit] 
      plot coordinates {
        (1,70.7) +- (0,1.1)
        (2,67.1) +- (0,1.2)
        (5,59.4) +- (0,1.3)
        (10,57.4) +- (0,1.4)
        (20,55.7) +- (0,1.5)
        (50,54.8) +- (0,1.4)
        (100,54.6) +- (0,1.4)
        (200,53.7) +- (0,1.4)
        (437,53.5) +- (0,1.5)
      };
      \addlegendentry{LRS3-TED (DNN-HMM)}

    \addplot[thick, dashed, mark=x, mark options={solid}, blue, error bars/.cd, y dir=both, y explicit] 
      plot coordinates {
        (1,100.0) +- (0,0.0)
        (2,100.0) +- (0,0.0)
        (5,98.0) +- (0,1.8)
        (10,67.2) +- (0,1.7)
        (20,58.8) +- (0,1.7)
        (50,53.1) +- (0,1.8)
        (100,50.9) +- (0,1.8)
        (200,50.3) +- (0,1.7)
        (437,50.0) +- (0,1.7)
      };
      \addlegendentry{LRS3-TED (CTC/Attention)}

    %% LRS2-BBC %%
    \addplot[thick, mark=x, mark options={solid}, black!80, error bars/.cd, y dir=both, y explicit] 
      plot coordinates {
        (1,38.4) +- (0,1.5)
        (2,35.3) +- (0,1.4)
        (5,31.9) +- (0,1.5)
        (10,31.5) +- (0,1.5)
        (20,31.5) +- (0,1.5)
        (50,31.1) +- (0,1.5)
        (100,30.2) +- (0,1.4)
        (223,29.8) +- (0,1.5)
      };
      \addlegendentry{LRS2-BBC (DNN-HMM)}

    \addplot[thick, dashed, mark=x, mark options={solid}, black!80, error bars/.cd, y dir=both, y explicit] 
      plot coordinates {
        (1,100.0) +- (0,0.0)
        (2,87.2) +- (0,1.3)
        (5,35.1) +- (0,1.6)
        (10,31.7) +- (0,1.6)
        (20,30.7) +- (0,1.5)
        (50,29.3) +- (0,1.5)
        (100,28.7) +- (0,1.4)
        (223,28.0) +- (0,1.5)
      };
      \addlegendentry{LRS2-BBC (CTC/Attention)}
      
    \end{axis}
  \end{tikzpicture}
\caption{Comparison in terms of performance (\% WER) of the DNN-HMM and the CTC/Attention decoders based on the number of hours used to estimate both paradigms. The 9, 223, and 437 hours refers to the entire training set of the LIP-RTVE, LRS2-BBC, and LRS3-TED databases, respectively.}
\label{fig:comparison}
\end{figure*}
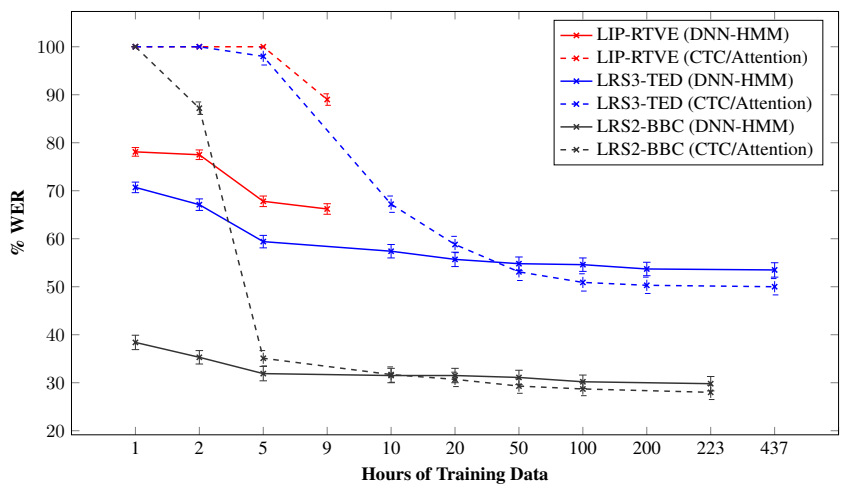

Experiments were conducted on a 12-core 3.50GHz Intel i7-7800X CPU and a GeForce RTX 2080 GPU with 8GB memory.

\subsection{Visual Speech Encoder} \label{sec:impl-encoder}

In most of our experiments, the visual speech encoder used the weights publicly released by \newcite{ma2022visual} for the LRS2-BBC database, where more than 1000 hours of data from different databases (including the LRS3-TED corpus) were considered. Only in the case of the LIP-RTVE database, due to language mismatch, the encoder was fine-tuned by assembling the LRS2-BBC encoder and its corresponding CTC/Attention decoder pre-trained with the weights publicly released by \newcite{ma2022visual} for the Spanish language. In order to represent the situation in a data-scarcity scenario, we only used 1 hour of data from the LIP-RTVE training set. Implementation details about this encoder fine-tuning process can be found in Subsection \ref{sec:training}.

\subsection{Transformer Language Model} \label{sec:impl-lm}

In our experiments, our Transformer-based LM (used both for the DNN-HMM lattice re-scoring and the CTC/Attention decoder) was pre-trained using the weights publicly released by Ma et al. \cite{ma2022visual} for both the English and Spanish language. Each of them was estimated using millions of characters collected from different databases corresponding to the language addressed. Nonetheless, it should be noted that, for the English LM, the transcriptions from the training sets of the LRS2-BBC and LRS3-TED databases were also considered. Therefore, to conduct a fair comparison, we fine-tuned the Spanish LM to the LIP-RTVE database using the 80k phrases provided by the original authors \cite{lrec2022liprtve}. Details on this LM fine-tuning process are described in Subsection \ref{sec:training}. Considering the same character vocabularies described in Subsection \ref{sec:ctc-attention}, our LMs presented a character-level perplexity of around 3.0 for the corresponding test set of all the proposed databases. Each LM comprises around 50 million parameters.
% 3.0, 2.7, and 3.4 for the LRS2-BBC, LRS3-TED and LIP-RTVE corpora, respectively.

\subsection{Training Process} \label{sec:training}

\noindent\textbf{Data Sets:} the official splits were kept, with slight variations for the English corpora. For the LRS2-BBC, the pre-training and training sets were condensed into one training set, comprising a total of 223 hours. Similarly, the `pre-train' and `train-val' sets from the LRS3-TED were used as a 437-hours training set. In both cases, utterances with more than 600 frames were excluded, as considered by \newcite{maja2021conformers,ma2022visual}.

\noindent\textbf{Conventional Hybrid Decoder.} Although we explored different training setups, the toolkit's default settings specified in Karel's  DNN-HMM implementation\footnote{\url{https://github.com/kaldi-asr/kaldi/blob/master/egs/wsj/s5/steps/nnet/train.sh}} provided the best recognition rates.

\noindent\textbf{CTC/Attention Decoder.} In all our experiments, we considered the settings specified by \newcite{ma2022visual}. Concretely, we used the Adam optimizer \cite{kingma2014adam} and the Noam scheduler \cite{vaswani2017attention} with 25,000 warmup steps during 50 epochs with a batch size of 16 samples, yielding a peak learning rate of 4$\times$10\textsuperscript{-4}. Regarding the CTC/Attention balance, we set the $\alpha$ weight of Equation \ref{eq:ctc-attention-loss} to 0.1.

\noindent\textbf{Fine-Tuning Settings.} The LM and the visual speech encoder were fine-tuned when addressing the LIP-RTVE database. In both cases, the AdamW optimizer \cite{loshchilov2017decoupled} and a linear one-cycle scheduler \cite{leslie2019onecycle} were used during 5 epochs, yielding a peak learning rate of 5$\times$10\textsuperscript{-5}. Due to our memory limitations, the batch size was set to 1 sample. We explored the accumulating gradient strategy \cite{ott2018accum}, but no significant differences were found, possibly because the normalization layers were still affected by the actual reduced batch size.

\subsection{Inference Process} \label{sec:inference}

\noindent\textbf{Conventional Hybrid Decoder.} As considered by \newcite{luscher19interspeech}, we applied an approach based on a combination of one-pass decoding and lattice re-scoring. First of all, with a beam size of 18, we explored word insertion penalties from -5.0 to 5.0 and LM scales from 1 to 20. Once the best setting was determined, a lattice composed of the best 100 hypothesis for each test sample was obtained using the 3-gram word-level LM. Afterwards, the lattice was re-scored using the Transformer-based LM. In all the cases, the visual speech decoder was scaled by a factor of 0.1.

\noindent\textbf{CTC/Attention Decoder.} As considered by \newcite{ma2022visual}, we incorporated the Transformer-based LM in a shallow fusion manner. According to Equation \ref{eq:ctc-attention-decoding}, we set the $\beta$ weight to 0.6 and 0.4 for English and Spanish, respectively. For English, we set a word insertion penalty of 0.5 and a beam size of 40, while for Spanish, we set the word insertion penalty and the beam size to 0.0 and 30, respectively. Regarding the CTC/Attention balance, we set the $\lambda$ weight of Equation \ref{eq:ctc-attention-decoding} to 0.1. It should be noted that we used the model averaged over the last 10 training epochs.

\noindent\textbf{Evaluation Metric:} Experiment results were reported in terms of the well-known Word Error Rate (WER) with 95\% confidence intervals using the method described by \newcite{bisani2004bootstrap}.

\section{Results \& Discussion}
\label{sec:results}

\begin{table}[!htbp]
\centering
\adjustbox{width=\columnwidth}{\begin{tabular}{cccccc}
\toprule
\multirow{2}{*}[-3pt]{\textbf{\begin{tabular}[c]{@{}c@{}}Training\\ Hours\end{tabular}}} & \multicolumn{2}{c}{\textbf{DNN-HMM}} & & \multicolumn{2}{c}{\textbf{CTC/Attention}} \\ \cmidrule{2-3} \cmidrule{5-6}
 & \textbf{\% WER} & \textbf{Time} & \textbf{} & \textbf{\% WER} & \textbf{Time} \\ \midrule
\textbf{1} & 38.4$\pm$1.5 & 3.9 &  & 100.0$\pm$0.0 & 3.3 \\
\textbf{2} & 35.3$\pm$1.4 & 6.3 &  & 87.2$\pm$1.3 & 5.8 \\
\textbf{5} & 31.9$\pm$1.5 & 9.4 &  & 35.1$\pm$1.6 & 15.8 \\
\textbf{10} & 31.5$\pm$1.5 & 15.9 &  & 31.7$\pm$1.6 & 30.8 \\
\textbf{20} & 31.5$\pm$1.5 & 25.9 &  & 30.7$\pm$1.5 & 61.7 \\
\textbf{50} & 31.1$\pm$1.5 & 58.4 &  & 29.3$\pm$1.5 & 151.7 \\
\textbf{100} & 30.2$\pm$1.4 & 114.0 &  & 28.7$\pm$1.4 & 306.7 \\
\textbf{223} & 29.8$\pm$1.5 & 221.7 &  & 28.0$\pm$1.5 & 634.2 \\ \bottomrule
\end{tabular}}
\caption{Comparison of the DNN-HMM and the CTC/Attention decoders for the LRS2-BBC database based on the number of hours used to estimate both paradigms. System performance (\% WER) and training time (Time) expressed in minutes are reported. The 223 training hours refer to the entire training set.}
\label{tab:lrs2}
\end{table}

\noindent\textbf{LRS2-BBC.} Using the LRS2-BBC database is the ideal scenario, where the proposed encoder extracts the visual speech features for which it was explicitly trained. Table \ref{tab:lrs2} reflects how the conventional paradigm is not only capable of obtaining state-of-the-art results, but also of significantly outperforming the CTC/Attention model in data-scarcity scenarios. Only when at least 10 hours of data were available, both approaches provided similar recognition rates. Besides, it should be mentioned that both paradigms presented a real time factor of around 0.75. However, despite offering a slightly better performance, the CTC/Attention estimation took more than twice the hybrid system training time.

\begin{table}[!htbp]
\centering
\adjustbox{width=\columnwidth}{\begin{tabular}{cccccc}
\toprule
\multirow{2}{*}[-3pt]{\textbf{\begin{tabular}[c]{@{}c@{}}Training\\ Hours\end{tabular}}} & \multicolumn{2}{c}{\textbf{DNN-HMM}} & \textbf{} & \multicolumn{2}{c}{\textbf{CTC/Attention}} \\ \cmidrule{2-3} \cmidrule{5-6}
 & \textbf{\% WER} & \textbf{Time} & \textbf{} & \textbf{\% WER} & \textbf{Time} \\ \midrule
\textbf{1} & 70.7$\pm$1.1 & 4.4 &  & 100.0$\pm$0.0 & 2.5 \\
\textbf{2} & 67.1$\pm$1.2 & 7.0 &  & 100.0$\pm$0.0 & 5.0 \\
\textbf{5} & 59.4$\pm$1.3 & 9.4 &  & 98.0$\pm$1.8 & 13.3 \\
\textbf{10} & 57.4$\pm$1.4 & 15.5 &  & 67.2$\pm$1.7 & 25.8 \\
\textbf{20} & 55.7$\pm$1.5 & 26.6 &  & 58.8$\pm$1.7 & 52.5 \\
\textbf{50} & 54.8$\pm$1.4 & 60.6 &  & 53.1$\pm$1.8 & 130.8 \\
\textbf{100} & 54.6$\pm$1.4 & 115.5 &  & 50.9$\pm$1.8 & 261.7 \\
\textbf{200} & 53.7$\pm$1.4 & 229.0 &  & 50.3$\pm$1.7 & 524.2 \\
\textbf{437} & 53.5$\pm$1.5 & 358.2 &  & 50.0$\pm$1.7 & 820.0 \\ \bottomrule
\end{tabular}}
\caption{Comparison of the DNN-HMM and the CTC/Attention decoders for the LRS3-TED database based on the number of hours used to estimate both paradigms. System performance (\% WER) and training time (Time) expressed in minutes are reported. The 437 training hours refer to the entire training set.}
\label{tab:lrs3}
\end{table}

\noindent\textbf{LRS3-TED.} It is considered our domain-mismatch scenario. First, it should be noted that this corpus was used during the pre-training stage of the visual speech encoder. However, due to the data-driven nature of the encoder and the fact that it was later fine-tuned to the LRS2-BBC database, the resulting features were expected to be worse than those extracted for the aforementioned corpus. Nonetheless, it allowed us to study whether the consequences of this deterioration in the quality of visual speech features could be mitigated when a larger amount of data is available. 

As Table \ref{tab:lrs3} reflects, results comparable to state of the art (25-30\% WER) were not achieved. Moreover, 20 hours were now necessary for both paradigms to offer a similar performance. The real time factor was around 0.82 and 0.75 for the DNN-HMM and CTC/Attention paradigm, respectively. Nonetheless, we can observe an analogous behaviour to that described for the LRS2-BBC database regarding data-scarcity scenarios, where the DNN-HMM would still be the best approach. Conversely, from 100 hours of data, the CTC/Attention showed significant differences w.r.t the conventional paradigm. It suggests that the state-of-the-art decoder could be more adaptable to poorer-quality speech representations.

However, the DNN-HMM and CTC/Attention decoders converge when the availability of more data does not imply any improvement in terms of performance (with 20 and 50 hours of training data, respectively, differences are not significant w.r.t. using all available data). This fact would demonstrate that the quality of the visual speech encoder is a real limitation whose consequences are not mitigated by decoders despite having larger amounts of data. 

\begin{table}[!htbp]
\centering
\begin{threeparttable}
\adjustbox{width=\columnwidth}{\begin{tabular}{cccccc}
\toprule
\multirow{2}{*}[-3pt]{\textbf{\begin{tabular}[c]{@{}c@{}}Training\\ Hours\end{tabular}}} & \multicolumn{2}{c}{\textbf{DNN-HMM}} & \textbf{} & \multicolumn{2}{c}{\textbf{CTC/Attention}} \\ \cmidrule{2-3} \cmidrule{5-6}
 & \textbf{\% WER} & \textbf{Time} & \textbf{} & \textbf{\% WER} & \textbf{Time} \\ \midrule
\textbf{1} & 78.1$\pm$0.9 & 3.5 &  & $>$100.0\tnote{$\dagger$} & 2.5 \\
\textbf{2} & 77.5$\pm$1.0 & 5.5 &  & $>$100.0\tnote{$\dagger$} & 5.0 \\
\textbf{5} & 67.8$\pm$1.1 & 9.9 &  & $>$100.0\tnote{$\dagger$} & 12.5 \\
\textbf{9} & 66.2$\pm$1.1 & 16.6 &  & 89.0$\pm$1.2 & 23.3 \\ \bottomrule
\end{tabular}}
\begin{tablenotes}
    \footnotesize
    \item[$\dagger$] due to a peculiarity of the WER metric
\end{tablenotes}
\end{threeparttable}
\caption{Comparison of the DNN-HMM and the CTC/Attention decoders for the LIP-RTVE database based on the number of hours used to estimate both paradigms. System performance (\% WER) and training time (Time) in minutes are reported. The 9 training hours refer to the entire training set.}
\label{tab:liprtve}
\end{table}

\noindent\textbf{LIP-RTVE.} In the case of the LIP-RTVE database, we were not only faced with a data-scarcity scenario, but also with a mismatch in terms of language. These could be the reasons why our first results were not acceptable. Therefore, we decided to adapt the visual speech encoder as if we were in the worst possible scenario of our experiments: when only 1 hour of data was available. As described in Subsection \ref{sec:visual-encoder}, we fine-tuned the encoder in an end-to-end manner, obtaining a baseline model capable of reaching results around 88.6\% WER. 

Results in Table \ref{tab:liprtve} show that, as in the rest of the studied scenarios, the conventional DNN-HMM decoder outperforms its CTC/Attention counterpart. Using around 10 hours of data w.r.t. only 1 hour enhances around 15\% WER in relative terms for the DNN-HMM system, which is in harmony with the roughly 18\% relative improvement observed for the LRS2-BBC and LRS3-TED corpora. Furthermore, we argue that one of the reasons that could be behind the success of the DNN-HMM paradigm was the word-level LM influence.

We also investigated fine-tuning the entire end-to-end architecture using the whole training set of the LIP-RTVE database. Recognition rates of around 60\% WER were obtained, which significantly improves the best performance obtained for the LIP-RTVE database to date. However, it should be noted that their estimate assumes more than six times the training time w.r.t the DNN-HMM model.

\noindent\textbf{Overall Analysis.} Figure \ref{fig:comparison} reflects how system performance degrades as visual speech features deteriorate from the ideal scenario (LRS2-BBC) to the language-mismatch (LIP-RTVE) scenario. This trend is not only an aspect we could expect, but it is also supported by Tseng et al. \cite{tseng2023av}, whose study demonstrated the lack of generalization of different audio-visual self-supervised speech representations in multiple tasks. However, the interesting finding is that the conventional DNN-HMM, compared to its state-of-the-art counterpart, offers a significantly more robust approach when the quality of our visual speech features is not optimal. If, for instance, we analyze the scenario with 5 hours of training data, we can observe how the performance gap between the DNN-HMM and CTC/Attention paradigms in the language- and domain-mismatch scenario is significantly greater than in ideal settings, making the DNN-HMM paradigm a more suitable option when addressing the task in data-scarcity scenarios and non-optimal visual speech features.

\noindent\textbf{Findings.} According to the findings of our case study, different aspects might be helpful for future researchers and developers focused on designing VSR systems in data-scarcity scenarios with limited computational resources. One of the main aspects is that, even in ideal scenarios with any type of limitation, DNN-HMM decoders not only reach state-of-the-art performance rates but also offer significantly lower training time costs. Furthermore, the fewer number of parameters composing it would facilitate the deployment of this type of system. Similarly, when we suffer from data scarcity and/or lack of optimal self-supervised speech representations for our specific conditions, the state-of-the-art CTC/Attention architecture would not be a recommendable option. 

\section{Conclusions \& Future Work}
\label{sec:conclusions}
In this work, we presented, to the best of our knowledge, the first thorough comparative study on the conventional DNN-HMM decoder and its state-of-the-art CTC/Attention counterpart for the visual speech recognition task. Unlike those comparative studies focused on auditory-based ASR, we also systematically investigated how these different decoding paradigms behave based on the amount of data available for their estimation. As reflected in Figure \ref{fig:comparison}, our results showed that the conventional approach achieved recognition rates comparable to the state of the art, significantly outperforming the CTC/Attention model in data-scarcity scenarios. In addition, the DNN-HMM approach offered valuable properties, such as reduced training time and fewer parameters. Finally, by exploring databases of different natures, experiments suggest that further research should still focus on improving the robustness of visual speech representations for data scarcity, as well as domain- and language-mismatch scenarios.

For this reason, one of our future lines of research is not only studying how state-of-the-art visual speech features can generalize to other tasks and domains, but also extending our work toward estimating and evaluating robust multi-lingual visual speech representations using the MuAViC benchmark \cite{anwar2023muavic}.

\section{Acknowledgements}
\label{sec:acknowledgements}
This work was partially supported by Grant CIACIF/2021/295 funded by Generalitat Valenciana and by Grant PID2021-124719OB-I00 under project LLEER (PID2021-124719OB-100) funded by MCIN/AEI/10.13039/501100011033/ and by ERDF, EU A way of making Europe.

\section{Bibliographical References}
\label{sec:bibliographical}
\bibliographystyle{lrec2022-bib}
\bibliography{lrec2022-example}

\end{document}